\documentclass{siamart190516}

\usepackage{lipsum}
\usepackage{amsfonts}
\usepackage{graphicx}
\usepackage{epstopdf}
\usepackage{algorithmic}
\ifpdf
  \DeclareGraphicsExtensions{.eps,.pdf,.png,.jpg}
\else
  \DeclareGraphicsExtensions{.eps}
\fi

\newsiamremark{remark}{Remark}
\newsiamremark{hypothesis}{Hypothesis}
\crefname{hypothesis}{Hypothesis}{Hypotheses}
\newsiamthm{claim}{Claim}

\headers{DEEP LEAST-SQUARES METHODS}{Z. Cai, J. Chen, M. Liu and X. Liu}

\title{DEEP LEAST-SQUARES METHODS: AN UNSUPERVISED LEARNING-BASED NUMERICAL METHOD FOR SOLVING ELLIPTIC PDEs\thanks{Submitted to the editors DATE.}}

\author{Zhiqiang Cai\thanks{Department of Mathematics, Purdue University, 150 N. University Street, West Lafayette, IN 47907-2067 
  (\email{caiz@purdue.edu}, \email{chen2042@purdue.edu},
  \email{liu1957@purdue.edu}).}
\and Jingshuang Chen\footnotemark[2]
\and Min Liu\thanks{School of Mechanical Engineering, Purdue University, 585 Purdue Mall,
West Lafayette, IN 47907-2088(\email{liu66@purdue.edu}) }
\and Xinyu Liu\footnotemark[2]}

\usepackage{amsopn}

\ifpdf
\hypersetup{
  pdftitle={FOSLS},
  pdfauthor={Z.Cai, J.Chen, M.Liu and X.Liu}
}
\fi

\usepackage[utf8]{inputenc}
\usepackage{bm}
\usepackage{mathtools}
\usepackage{indentfirst}
\usepackage{subfigure}
\usepackage{tcolorbox}
\usepackage{color}
\usepackage[export]{adjustbox}

\usepackage{algorithmic}
\Crefname{ALC@unique}{Line}{Lines}

\newcommand{\R}{\mathbb{R}}

\usepackage{graphicx}
\usepackage{diagbox}
\usepackage{pifont}
\newcommand{\vertiii}[1]{{\left\vert\kern-0.25ex\left\vert\kern-0.25ex\left\vert #1 
    \right\vert\kern-0.25ex\right\vert\kern-0.25ex\right\vert}}

\newcommand{\bsigma}{\mbox{\boldmath${\sigma}$}}

\newcommand{\btau}{\mbox{\boldmath${\tau}$}}

\begin{document}

\maketitle
\begin{abstract}
This paper studies an unsupervised deep learning-based numerical approach for solving partial differential equations (PDEs). The approach makes use of the deep neural network to approximate solutions of PDEs through the compositional construction and employs least-squares functionals as loss functions to determine parameters of the deep neural network. 
There are various least-squares functionals for a partial differential equation. This paper focuses on 
the so-called first-order system least-squares (FOSLS) functional
studied in \cite{cai1994first}, which is based on a first-order system of scalar second-order elliptic PDEs. Numerical results for second-order elliptic PDEs in one dimension are presented.
\end{abstract}



\section{Introduction}
Recently, deep neural network (DNN) models have had great success in computer 
vision, pattern recognition, and many other artificial intelligence tasks.  
A special feature of DNN is its new way to approximate functions through a composition of multiple linear and activation functions. This leads to some recent studies (see, e.g., \cite{Berg18,Tim19,Weinan18, Sirignano18}) on applications of deep learning to partial differential equations (PDEs). 

The idea of solving differential equations using neural networks may be traced back to a paper in 1994 by Dissanayake and Phan-Thien \cite{Dissanayake94}. 
For a differential equation $L(u) = 0$ defined on the domain $\Omega$ with boundary condition $B(u) = 0$
on $\partial\Omega$, a neural network was trained to minimize the following least-square functional
\begin{equation}\label{ls_functional}
\tilde{\mathcal{L}}(v)=\int_\Omega{\Big|L(v)(x)\Big|^2\,dx} + \int_{\partial\Omega}{\Big|B(v)(x)\Big|^2ds}\,
\equiv \|L(v)\|^2_{0,\Omega} + \|B(v)\|^2_{0,\partial\Omega},
\end{equation}
where $\|\cdot\|_{0,S}$ is the $L^2$ norm over subdomain $S=\Omega$ or $\partial\Omega$.
Several follow-up works use similar ideas with one hidden layer and sampling points from a mesh to numerically approximate the integrals in $\tilde{\mathcal{L}}$ at each iteration \cite{Lagaris98,Lagaris00,McFall09}. More recently, there is a limited emerging literature on the use of deeper hidden layers to solve PDEs \cite{Berg18,Tim19,Sirignano18}. It is also illustrated that the sampling points can be obtained by a random sampling of the domain rather than using a mesh, which is beneficial in higher-dimensional problem \cite{Berg18,Sirignano18}. 
The least-squares functional defined in (\ref{ls_functional}) is based on the original PDEs. For a second order PDE, the minimization of 
$\tilde{\mathcal{L}}(v)$ over admissible functions leads
to a fourth-order PDE, which is a more difficult problem than the original one. Moreover, the interior and the boundary integrals in (\ref{ls_functional}) are not balanced.

Another formulation of the loss function is to use the energy functional of the underlying PDEs, such as the resulting deep Ritz method recently introduced by E-Yu \cite{Weinan18}. For a Poisson problem with Dirichlet boundary conditions, i.e.,
\begin{equation*}
\left\{ \begin{array}{rclll}
 -\Delta u &=&f, \quad & \mbox{in} & \Omega,\\[2mm]
 u&=&0, \quad & \mbox{on}& \partial \Omega,
 \end{array}\right.
\end{equation*}
the energy functional is given by
\begin{equation}\label{related3}
\tilde{\mathcal{J}}(v)=\int_\Omega{\left(\frac{1}{2}|\nabla v(x)|^2-f(x)v(x)\right)dx}.
\end{equation}
This approach is applicable to problems having an underlying minimization principle.

The purpose of this paper is to study an unsupervised deep learning-based numerical approach for solving PDEs. The approach makes use of a deep neural network to approximate solutions of PDEs through the compositional construction and employs least-squares (LS) functionals
as loss functions to determine parameters of the deep neural network. 
There are various least-squares functionals for a partial differential equation, 
this paper focuses on the FOSLS functional
studied in \cite{cai1994first}, which is based on a first-order system of scalar second-order elliptic PDEs.

The LS methodology has been intensively studied for many PDEs including problems arising from
solid and fluid dynamics, radiation transport, magnetohydrodynamics, etc. 
The method has many attractions. The two striking features are (i) it
naturally symmetrizes and stabilizes the original problem; 
and (ii) value of the corresponding LS functional at the current approximation is an accurate a posteriori error estimator.
The first property enables us to work on complex systems which might not have underlying minimization principles, and the second one provides feedback for automatically controlling numerical processes
such as the number and the location of quadrature points for evaluating LS functional.

The paper is organized as follows. Section 2 describes the second order elliptic PDEs, 
the least-squares formulation based on a first-order system of the underlying problem introduced in \cite{cai1994first},
and proper treatment of boundary conditions when using energy, LS, and FOSLS functionals. Section 3 introduces 
deep neural network and corresponding deep FOSLS method. Finally, numerical results on three test problems in one dimension are presented in section 4. Moreover, a numerical comparison between uniformly distributed and adaptively obtained quadrature points is reported in section 4.4.

\section{Problem Formulation}
Let $\Omega$ be a bounded domain in ${\R}^d$ with
Lipschitz boundary $\partial\Omega= \bar{\Gamma}_D \cup \bar{\Gamma}_N$. Consider the following second-order scalar elliptic partial
differential equation:
 \begin{equation}\label{pde}
 -\mbox{div}\, (A \nabla \,u) +Xu = f, \quad \mbox{in }  \Omega\subset \R^d
\end{equation}
with boundary conditions
 \begin{equation}\label{bc1}
  u=  g_{_{\small D}}, \quad  \mbox{on} \ \Gamma_D
  \quad\mbox{and}\quad
  -{\bf n} \cdot A \nabla \, u = g_{_{\small N}}, \quad  \mbox{on} \ \Gamma_N,
\end{equation}
where $f \in L^2(\Omega)$, $g_{_{\small D}}\in H^{1/2}(\Gamma_D)$, $g_{_{\small N}}\in H^{-1/2}(\Gamma_N)$;
$A(x)$ is a $d \times d$ symmetric
matrix-valued function in $L^2(\Omega)^{d\times d}$; $X$ is a linear differential
operator of order at most one; 
and ${{\bf n}}$ is the outward
unit vector normal to the boundary. We assume that $A$ is uniformly
positive definite. Possible choices for $X$ include:
$Xu= \mbox{div} \,({\bf b} \,u)$ with ${\bf b}  \in L^2(\Omega)^d$ and $Xu = {\bf a}\cdot \nabla \, u + c u$
with ${\bf a} \in L^2(\Omega)^d, ~c(x) \in L^2(\Omega)$.

Here and thereafter, we use the standard notation and definitions for the Sobolev space $H^{s}(\Omega)$ and $H^{s}(\Gamma)$ for a subset $\Gamma$ in $\partial \Omega$. 
The standard associated inner product and norms are denoted by $(\cdot,\cdot)_{s,\Omega}$ and $(\cdot,\cdot)_{s,\Gamma}$ and by $\Vert \cdot \Vert _{s,\Omega}$ and $\Vert \cdot \Vert _{s,\Gamma} $, respectively. 
When $s=0$, $H^{0}(\Omega)$ coincides with $L^2(\Omega)$.
Denote the corresponding norms on product space $H^s(\Omega)^d$ by $\|\cdot\|_{s,\,\Omega,\,d}$ and
$|\cdot|_{s,\,\Omega,\,d}$.  When there is no ambiguity, the subscript $\Omega$ and $d$ in the designation of norms  will be suppressed.

\subsection{Least-Squares Formulations}

Problem (\ref{pde})-(\ref{bc1}) is non-symmetric in general and, hence, has  no underlying minimization principle. 
To make use
of the deep neural network, we will employ LS principles.  There are many LS formulations for problem (\ref{pde}). For example, a direct application of the LS principle to problem (\ref{pde}) leads to
a LS functional defined in (\ref{ls1a}) which is similar to that in (\ref{ls_functional}) but with different boundary terms.
In this section, we describe the FOSLS formulation introduced in \cite{cai1994first} which is based on a first-order
system of problem (\ref{pde})-(\ref{bc1}).

To this end, introducing the flux variable $\bsigma = -A\nabla u$, the second-order problem in (\ref{pde}) may be
rewritten as a first-order system:
\begin{equation}\label{fos}
\left\{ \begin{array}{rclll}
 \mbox{div}\, \bsigma +Xu & = & f, & \mbox{in} & \Omega,\\[2mm]
 \bsigma + A\nabla u &=& {\bf 0}, & \mbox{in} & \Omega
 \end{array}\right.
 \end{equation}
with boundary conditions
  \begin{equation}\label{bc2}
  u= g_{_{\small D}}, \quad  \mbox{on} \ \Gamma_D
  \quad\mbox{and}\quad
  {\bf n} \cdot \bsigma = g_{_{\small N}}, \quad  \mbox{on} \ \Gamma_N.
\end{equation} 
Let 
$$
H(\mbox{div};\Omega) \equiv  \left\{ {\bf v} \in L^2(\Omega)^d:
\; \,\mbox{div} \, {\bf v} \in L^2(\Omega).
\right\}.
$$
Denote subsets of $H^1(\Omega)$ and $H(\mbox{div};\Omega)$ satisfying non-homogeneous boundary conditions by 
 \[
 H^1_{_{\small D},g}(\Omega)=\{v\in H^1(\Omega) : v|_{\Gamma_D}=g_{_{\small D}}\}
 \text{ and }
 H_{_{\small N},g}=\{\btau\in H(\mbox{div};\Omega) : \btau\cdot {\bf n}|_{\Gamma_N}=g_{_{\small N}}\}
 \]
respectively. When $g_{_{\small D}}=0$ and $g_{_{\small N}}=0$, these subsets become subspaces and are denoted by $H^1_{_{\small D}}(\Omega)$ and $H_{_{\small N}}(\mbox{div};\Omega)$. Let 
 \[
 \mathcal{V}_g= H_{_{\small N},g}(\mbox{div};\Omega)\times H^1_{_{\small D},g}(\Omega)
 \quad\text{and}\quad
 \mathcal{V}_0= H_{_{\small N}}(\mbox{div};\Omega)\times H^1_{_{\small D}}(\Omega),
 \]
then the FOSLS formulation is to find $(\bsigma,\, u)\in \mathcal{V}_g$ such that
 \begin{equation}\label{fosls}
 \tilde{\mathcal{G}}(\bsigma,\,u;{\bf f}) = \min_{\small (\btau, v)\in \mathcal{V}_g} \tilde{\mathcal{G}}(\btau,\,v;{\bf f}),
\end{equation}
where ${\bf f}=(f,g_{_{\small D}},g_{_{\small N}})$ and the FOSLS functional is defined by
 \begin{equation}\label{fosls1}
 \tilde{\mathcal{G}}(\btau,\,v;{\bf f}) = \|\mbox{div}\, \btau+Xv -f\|^2_{\small 0,\Omega}+\|A^{-1/2}\btau + A^{1/2}\nabla v\|^2_{\small 0,\Omega}.
  \end{equation}

It has been proved in \cite{cai1994first} that the homogeneous 
FOSLS functional $\tilde{\mathcal{G}}(\btau,\,v;{\bf 0})$ is coercive and bounded in $\mathcal{V}_0$, i.e., there exist positive constants $c_1$ and $c_2$ such that
 \begin{equation}\label{ellipticity}
 c_1 \vertiii{(\btau,\,v)}^2 \leq \tilde{\mathcal{G}}(\btau,\,v;{\bf 0})  \leq c_2 \vertiii{(\btau,\,v)}^2
 \end{equation}
for all $(\btau,\,v)\in \mathcal{V}_0$, where the FOSLS energy norm is given by  
 \[\vertiii{(\btau,\,v)} = \left(
  \|\btau\|^2_{\small 0,\Omega}+\|\mbox{div}\,\btau\|^2_{\small 0,\Omega} + \|v\|^2_{\small 1,\Omega} \right)^{1/2}.
  \]
The corcevity and boundedness of the homogeneous FOSLS functional further implies that the FOSLS minimization problem in (\ref{fosls}) is well-posed, i.e., (\ref{fosls}) has a unique solution (see \cite{cai1994first}
for a detail discussion).

\subsection{Treatment of Boundary Conditions}

Unlike finite element functions, it is not easy for a deep neural network function to satisfy a prescribed boundary condition. Such a difficulty was observed in \cite{Weinan18} for the deep Ritz method. To circumvent this obstacle, for a Poisson equation (i.e., $A = I$ and $X =0$) with pure Dirichlet boundary conditions (i.e, $\Gamma_{_{\small N}}=\emptyset$), 
they add the essential boundary conditions to the energy functional:
\begin{equation}\label{energy1}
\tilde{\mathcal{J}}(v)=\int_\Omega{\left(\frac{1}{2}|\nabla v(x)|^2-f(x)v(x)\right)dx} +\beta\,\|v(x)-g_{_{\small D}}\|^2_{0,\partial\Omega},
\end{equation}
where $\beta$ is a parameter to be determined. When the data vanishes, i.e., $f=0$ and $g_{_{\small D}}=0$, the modified energy functional becomes
 \[
 \tilde{\mathcal{J}}(v)
 = \dfrac12\,\|\nabla v\|^2_{0,\Omega} +\beta\|v(x)\|^2_{0,\partial\Omega}.
 \]
By the Sobolev trace theorem, the interior and boundary norms in the above formula are not in the same scale. Specifically, the boundary norm is $1/2$-order weaker than the interior norm.
This consideration suggests the following modified energy functional of (2.8)
\begin{equation}\label{energy2}
\mathcal{J}(v;{\bf f})=\int_\Omega{\left(\frac{1}{2}|\nabla v(x)|^2-f(x)v(x)\right)dx}
+\beta\,\|v(x)-g_{_{\small D}}\|^2_{1/2,{\small \partial \Omega}},
\end{equation}
where ${\bf f}=(f,g_{_{\small D}})$ and $\beta$ is a constant. For the Poisson equation with the mixed boundary conditions in (\ref{bc1}), the energy functional becomes
\begin{equation}\label{energy}
\mathcal{J}(v;{\bf f})=\dfrac12\,\|\nabla v\|^2_{0,\Omega} - \left(\int_\Omega f(x)v(x)\,dx +\int_{\Gamma_N}g_{_{\small N}}v\,dS\right)
+\beta\,\|v(x)-g_{_{\small D}}\|^2_{1/2,{\small \Gamma_D}}
\end{equation}
where ${\bf f}=(f,g_{_{\small D}},g_{_{\small N}})$ and $\beta$ is a constant. The minimization problem based on the above energy functional is to find $u\in H^1(\Omega)$ such that
 \begin{equation}\label{mini}
 \mathcal{J}(u;\,{\bf f})=\min_{v\in H^1(\Omega)} \mathcal{J}(v;\,{\bf f}).
 \end{equation}

For the FOSLS formulation defined in (\ref{fosls}), both the Dirichlet and Neumann boundary conditions are essential boundary conditions and, hence, we need to add them to the FOSLS functional with proper scales:
 \begin{eqnarray}\label{fosls1a}\nonumber
 \mathcal{G}(\btau,\,v;{\bf f}) 
 &=& \|\mbox{div}\, \btau+Xv -f\|^2_{\small 0,\Omega}+\|A^{-1/2}\btau + A^{1/2}\nabla v\|^2_{\small 0,\Omega}  \\[2mm]
 && \quad  + \alpha_{\small D}\|v-g_{_{\small D}}\|^2_{1/2, \small\Gamma_D} + \alpha_{\small N} \|{\bf n} \cdot \btau - g_{_{\small N}}\|^2_{-1/2,\small\Gamma_N}
  \end{eqnarray}
for all $(\btau,\, v)\in \mathcal{V}\equiv H(\mbox{div};\Omega)\times H^1(\Omega)$, where $\alpha_{\small D}$
and $\alpha_{\small N}$ 
are constants and may be chosen to be one. Now, the corresponding FOSLS formulation is to find $(\bsigma,\, u)\in \mathcal{V}$ such that
 \begin{equation}\label{fosls_a}
 \mathcal{G}(\bsigma,\,u;{\bf f}) = \min_{\small (\btau, v)\in \mathcal{V}} \mathcal{G}(\btau,\,v;{\bf f}).
\end{equation}
It has been proved that the homogeneous 
FOSLS functional $\mathcal{G}(\btau,\,v;{\bf 0})$ is coercive and bounded in $\mathcal{V}$. This in turn implies that the LS minimization problem in (\ref{fosls_a}) is well-posed in the space $\mathcal{V}$ without strongly enforced boundary conditions.

For the LS functional defined in (\ref{ls_functional}), the norm on boundary conditions is weaker than that for the equation; moreover, the Dirichlet and the Neumann boundary conditions are not treated differently. A balanced LS functional for problem (\ref{pde}) is as follows:
 \begin{equation}\label{ls1a}
 \mathcal{L}(v;\,{\bf f}) = \| -\mbox{div}\, (A \nabla \,v) +Xv - f\|^2_{0,\Omega} + \beta_{\small D}\|v-g_{_{\small D}}\|^2_{3/2, \small\Gamma_D} + \beta_{\small N} \|{\bf n} \cdot A \nabla \, v + g_{_{\small N}}\|^2_{1/2,\small\Gamma_N},
 \end{equation}
where ${\bf f}=(f,g_{_{\small D}},g_{_{\small N}})$.
Now, the corresponding LS formulation is to find $u\in H^2(\Omega)$ such that
 \begin{equation}\label{ls_a}
 \mathcal{L}(u;\,{\bf f})  = \min_{\small v\in H^2(\Omega)}  \mathcal{L}(v;\,{\bf f}) .
 \end{equation}
Assume that the solution of problem (\ref{pde})-(\ref{bc1}) is $H^2$ regular. Then it is a direct consequence that the homogeneous 
LS functional $\mathcal{L}(v;\,{\bf 0})$ is coercive and bounded in $H^2(\Omega)$. This implies that problem (\ref{ls_a}) is well-posed by Lax-Milgram theorem \cite{cai1994first}.
 
 \begin{remark}
Note that the LS formulation (\ref{ls1a})-(\ref{ls_a}) is only applicable to problems whose solutions are sufficiently smooth, more precisely, at least in $H^2(\Omega)$. 
This, in turn, implies that a DNN with non-piecewise-linear activation function is needed when using the LS 
functional as the loss function.
 \end{remark}

\section{The Deep FOSLS}

This section describes deep neural network structures and the deep FOSLS method. Discussions on 
numerical evaluation of the FOSLS functional  are, in principle, valid for both the energy and the LS functionals.
Moreover, similar error bounds in (\ref{error}) and (\ref{error+alg}) for the deep FOSLS is also valid for the energy and the LS functionals
in the respective $H^1$ and $H^2$ norms.

\subsection{Deep Neural Network Structure}
\begin{figure}[ht]
\centering
    \includegraphics[width=4in, height=2.2in]{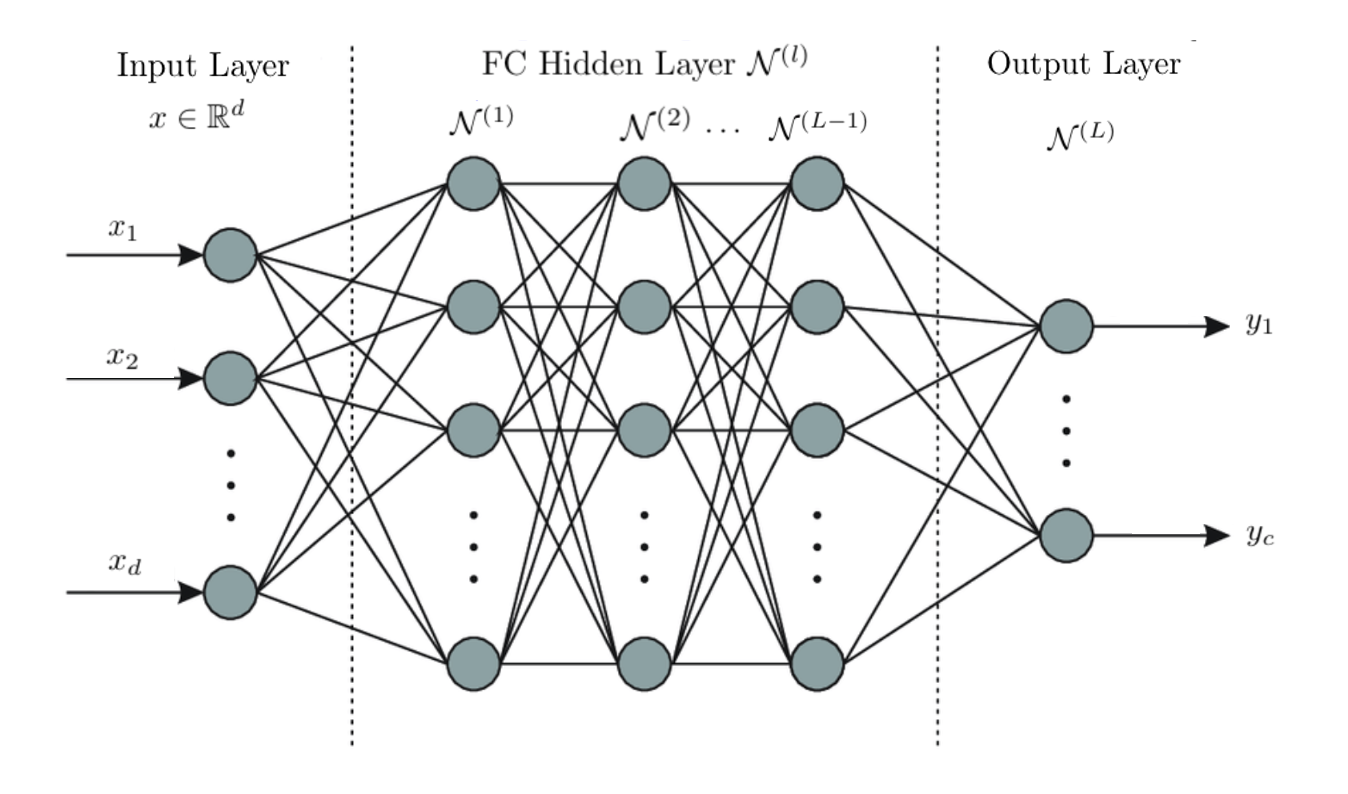}
    \caption{Fully-Connected Neural Network}
    \label{neural_network}
\end{figure}

For convenience of audiences in numerical analysis, in this section we describe the DNN structure through functional terminology. 
A deep neural network defines a function
\[
\mathcal{N}:\, x\in\R^{d}
\longrightarrow y=\mathcal{N}(x)\in\R^{c},
\]
where $d$ and $c$ are dimensions of input $x\in \R^d$ and output $y=\mathcal{N}(x)\in \R^c$, respectively. 
The DNN function $\mathcal{N}(x)$ is typically represented as compositions of many different layers of functions:  
\begin{equation}\label{DNN}
 y=\mathcal{N}(x)=\mathcal{N}^{(L)} \circ \cdots \mathcal{N}^{(2)}\circ \mathcal{N}^{(1)}(x),
\end{equation}
where the symbol $\circ$ denotes the composition of functions: $f\circ g(x) = f(g(x))$, and $L$ is the depth of the network. In this case, $\mathcal{N}^{(1)}$ is called the first layer of the network, $\mathcal{N}^{(2)}$ is called the second layer, and so on. All layers except the last one $\mathcal{N}^{(L)}$ are called hidden layers since they are hidden in between input and output (See Figure \ref{neural_network}).

Each layer is typically a vector-valued function. The choice of the function  $\mathcal{N}^{(l)}(x)$ is guided by many mathematical and engineering disciplines. In this paper, we use fully connected (FC) hidden layers. A FC layer $\mathcal{N}^{(l)}: \R^{n_{l-1}} \rightarrow \R^{n_{l}}$ is defined as a composition of a linear transformation $T^l: \R^{n_{l-1}} \rightarrow \R^{n_{l}}$ and an activation function $\psi^l: \R \rightarrow \R$ as follows:
\begin{equation}\label{layerdef}
   \mathcal{N}^{(l)}(x^{l-1})=\psi^l \circ T^l(x^{l-1})= \psi^l ({W}^lx^{l-1}+b^l),
   \quad\mbox{for } x^{l-1}\in \R^{n_{l-1}},
\end{equation}
where $W^l=\left(w^l_{ij}\right)_{n_{{l}}\times n_{l-1}}\in \R^{n_{l}\times n_{l-1}}$, $b^l\in \R^{n_{l}}$, and application of $\psi^l$ to a vector $z\in \R^{n_{l}}$ is defined component-wisely,  i.e., $\psi^l(z)=\left(\psi^l(z_i)\right)_{n_{{l}}\times 1}$. Components of $W^l$ and $b^l$ are called weights and bias, respectively, and are parameters to be determined (trained). 
Each component of the vector-valued function $\mathcal{N}^{(l)}$ is interpreted as a neuron and the dimensionality $n_{l}$ defines the width or the number of neurons of the $l^{\text{th}}$ layer in a network.
The $n_0=d$ and $n_L=c$ are the respective dimensions of input and output.
There are $n_{l}\times (n_{l-1}+1)$ parameters at the $l^{\text{th}}$ layer, and the total number of parameters of the DNN function $\mathcal{N}(x)$ defined in (\ref{DNN}) is
given by 
 \[
 N= \sum^L_{l=1} n_{l}\times (n_{l-1}+1).
 \]
 
Choices of the activation function $\psi$ have influences on the output of a model, its accuracy, and the computational efficiency of training. A commonly used activation function is the leaky ReLU defined as follows:
\begin{equation}\label{ReLU}
\psi(x)=\left\{\begin{array}{rclll}
 x, & \mbox{if }  x>0,\\[2mm]
 0.01x, & \mbox{otherwise,}
 \end{array}\right.
 \end{equation}
which is a continuous piecewise linear function. A DNN with a piecewise linear activation function is capable of generating rich function classes. For instance, as discussed in \cite{arora2018,tarela1999region}, a DNN with 
at most $[\log_2 (d+1)]$ hidden layers can represent piecewise linear function $\R^d \rightarrow \R$. Furthermore, by introducing some special network structures and adding more neurons as well as layers, DNN is able to approximate a large class of functions other than linear \cite{yarotsky2017error}. 

\begin{figure}[ht]
\centering
\includegraphics[width=3in,height=2in]{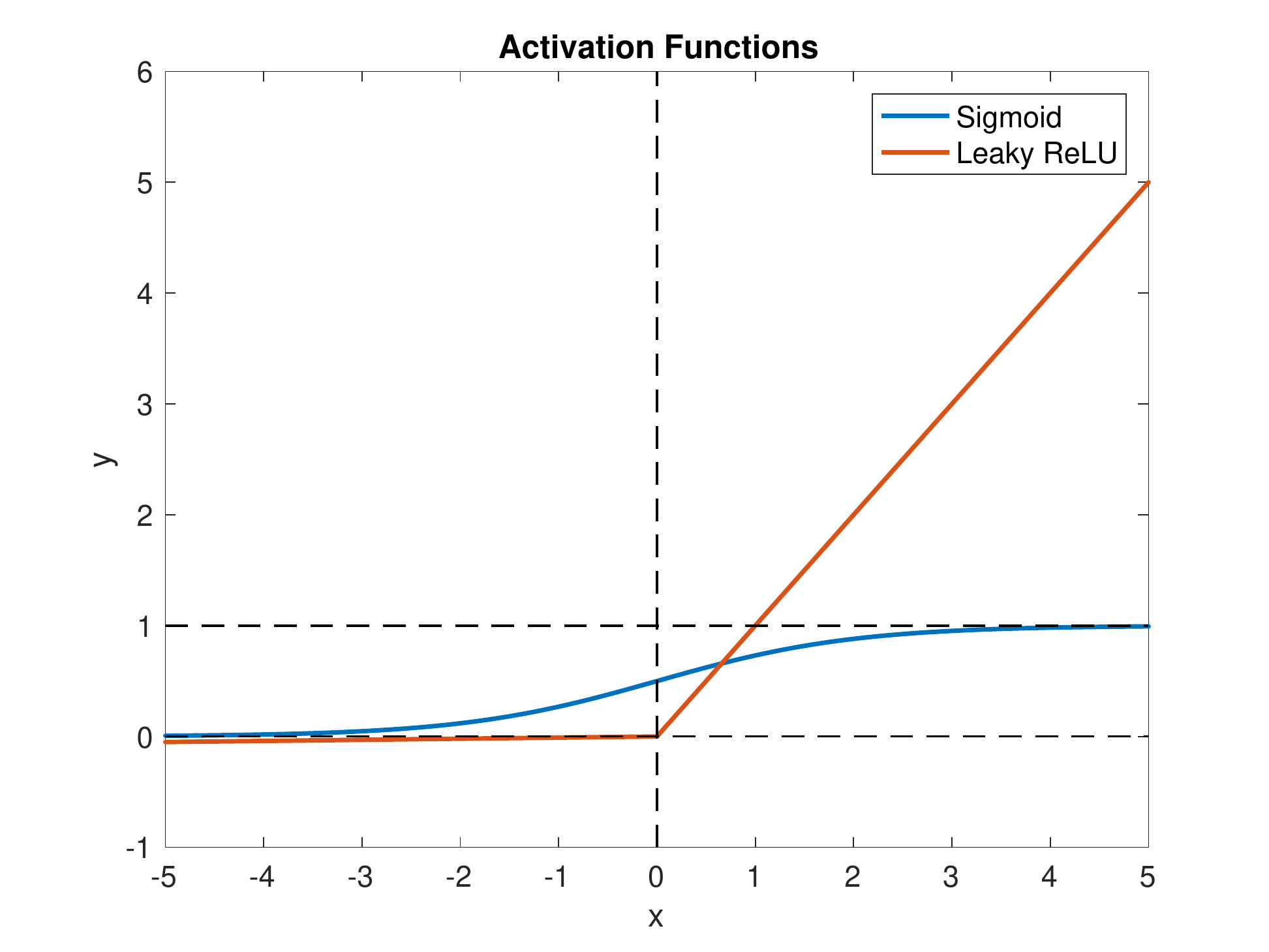}
\caption{Activation functions}\label{sigmoidpic}
\end{figure}

The sigmoid function is another commonly used activation function, which is defined by
\begin{equation}\label{sigmoid}
    \psi(x)=\dfrac{1}{1+e^{-x}}, \quad x\in \R.
\end{equation}
Both the leaky ReLU and the sigmoid activation functions 
are depicted in Figure \ref{sigmoidpic}.
The leaky ReLU is easier to compute than the non-linear sigmoid function. But using a smooth activation function such as the sigmoid function is essential for the deep LS method based on the
LS functional defined in either (\ref{ls_functional}) or (\ref{ls1a}). This is because functions generated by a DNN with a continuous piecewise linear activation function is only in $H^1(\Omega)$.

\subsection{Deep FOSLS}
The idea of the deep FOSLS is to employ DNN functions for approximating the solution $(\bsigma(x),\,u(x))$ of the FOSLS minimization problem in (\ref{fosls}). More specifically, for each $x\in \Omega\subset \R^d$, a DNN is implemented to compute an approximation $(\hat{\bsigma}(x, \Theta),\,\hat{u}(x, \Theta))$ at the point $x$, where $\Theta\in \R^N$ stands for all parameters (weights and biases) in the DNN. A deep FOSLS approximation is to find $(\hat{\bsigma}(x, \Theta),\,\hat{u}(x, \Theta))$ such that
 \begin{equation}\label{deep-fosls}
 \mathcal{G}(\hat{\bsigma}(x, \Theta),\,\hat{u}(x, \Theta);{\bf f}) 
 = \min_{\small \tilde{\Theta}\in \R^N} \mathcal{G}(\hat{\btau}(x, \tilde{\Theta}),\,\hat{v}(x, \tilde{\Theta});{\bf f}).
\end{equation}

Instead of evaluating the FOSLS functional analytically, in this paper we consider numerical approximation to the FOSLS functional.
This means that we will use numerical quadrature to approximate integrals of the FOSLS functional. For simplicity and generality in high dimensions, we will adopt composite ``mid-point'' quadrature rule.
To this end, let 
\[
{\cal T}=\{K\, :\, K\mbox{ is an open subdomain of } \Omega\}
\]
be a partition of the domain $\Omega$. Here, the partition means that union of all subdomains of ${\cal T}$ equal the whole domain $\Omega$ and that any two distinct subdomains of ${\cal T}$ have no intersection; more precisely,
 \[
 \bar{\Omega} = \cup_{K\in {\cal T}} \bar{K}
 \quad\mbox{and}\quad
 K\cap T = \emptyset,
 \quad \forall\,\, K,\, T \in {\cal T}.
 \]
Denote by ${\cal E}_{\small D}=\{E\, :\, E
=\partial K \cap \Gamma_D,\,\, \forall\, K\in\mathcal{T}\}$ 
and ${\cal E}_{\small N}=\{K\, :\, E=\partial K \cap \Gamma_N,\,\, \forall\, K\in\mathcal{T} \}$ 
the partitions of $\Gamma_D$ and $\Gamma_N$ associated with the partition $\mathcal{T}$, respectively.
Let $x_K$ and $x_E$ be interior points of $K\in {\cal T}$ and $E\in {\cal E}_S$ with $S=D$ or $N$, respectively. 
The $x_K$ and $x_E$ will be used as quadrature points below. Note that quadrature points are fundamentally different from sampling points used in the setting of supervised learning.
 
Since Sobolev norms $\|\cdot\|_{1/2}$ and $\|\cdot\|_{-1/2}$ in the FOSLS functional are not computationally feasible, we will approximate them by weighted $L^2$ norms with local weights $h^{-1/2}_E$ and $h^{1/2}_E$, respectively, where $h_E$ is the diameter of $E$. This idea leads to the following discrete FOSLS functional:
   \begin{eqnarray}\nonumber   \label{d-functiona11l}
   \hat{\mathcal{G}} (\hat{\btau}(x, \Theta),\,\hat{v}(x, \Theta);{\bf f}) 
   &&= \sum_{K\in {\cal T}} \left( \big(\mbox{div}\, \hat{\btau}+X\hat{v} -f\big)^2 (x_{\small K}, \Theta)
    +\big( A^{-1/2} \hat{\btau} + A^{1/2}\nabla \hat{v}  \big)^2 (x_{\small K}, \Theta)\right) |K| \\
   \label{d-functional}
   &&+\alpha_{_{\small D}}\sum_{E\in {\cal E}_D} \big( \hat{v}-g_{_{\small D}} \big)^2 (x_E, \Theta) |E|\,h^{-1}_E
   + \alpha_{_{\small N}}\sum_{E\in {\cal E}_N} \big( {\bf n} \cdot \hat{\btau} - g_{_{\small N}}\big)^2 (x_E, \Theta) |E|\,h_E,
   \end{eqnarray}
where $|K|$ and $|E|$ are the $d$ and $d-1$ dimensional measures of $K$ and $E$ respectively; and $\alpha_{_{\small D}}$ and $\alpha_{_{\small N}}$ are two positive constants. 
For given data $f$, $g_{\small D}$, and $g_{\small N}$, the value of the discrete FOSLS functional at $(\hat{\btau},\,\hat{v})$ is a function of the parameters $\Theta$.
Then the discrete deep FOSLS approximation is 
to find $(\hat{\bsigma}_{_{\small {\cal T}}}(x, \Theta),\,\hat{u}_{_{\small {\cal T}}}(x, \Theta))$ such that
 \begin{equation}\label{d-deep-fosls}
  \hat{\mathcal{G}} (\hat{\bsigma}_{_{\small {\cal T}}}(x, \Theta),\,\hat{u}_{_{\small {\cal T}}}(x, \Theta);{\bf f}) 
 = \min_{\small \tilde{\Theta}\in \R^N} \hat{\mathcal{G}} (\hat{\btau}
 (x, \tilde{\Theta}),\,\hat{v}
 (x, \tilde{\Theta});{\bf f}).
\end{equation}

\begin{remark}
Similar to the discrete FOSLS functional defined in {\it (\ref{d-functiona11l})}, the discrete energy and the discrete LS functionals are defined as follows:
\begin{eqnarray*}
\hat{\mathcal{J}}(\hat{v}(x,\Theta_u);{\bf f})
 &=&  \sum_{K\in {\cal T}} \left(\dfrac12\,|\nabla \hat{v}|^2 -  f\hat{v}\right)(x_K,\Theta_u)|K|
 - \sum_{E\in {\cal E}_N} \big(g_{_{\small N}}\hat{v} \big) (x_E, \Theta_u) |E| \\[2mm]
 && \quad
   +\, \alpha_{_{\small D}}\sum_{E\in {\cal E}_D} \big( \hat{v} - g_{_{\small D}}\big)^2 (x_E, \Theta_u) |E|\,h^{-1}_E \\[2mm]
\mbox{and }\,
\hat{\mathcal{L}}(\hat{v}(x,\Theta_u);{\bf f})
 &=& \sum_{K\in {\cal T}} \left(-\mbox{div}\, (A \nabla \,\hat{v}) +X\hat{v} - f\right)^2(x_K,\Theta_u)|K| \\[2mm]
 &+&\alpha_{_{\small D}}\!\!\sum_{E\in {\cal E}_D} \big( \hat{v}-g_{_{\small D}} \big)^2 (x_E, \Theta_u) |E|h^{-3}_E
   + \alpha_{_{\small N}}\!\!\sum_{E\in {\cal E}_N} \big( {\bf n} \cdot A\nabla \hat{v} + g_{_{\small N}}\big)^2 (x_E, \Theta_u) |E|h^{-1}_E,
\end{eqnarray*}
respectively, where $\alpha_D$ and $\alpha_N$ are positive constants.
\end{remark}

To understand approximation property of the discrete deep FOSLS method, by the triangle inequality, we have
\begin{equation}\label{error}
    \vertiii{ (\bsigma-\hat{\bsigma}_{_{\small {\cal T}}}, u - \hat{u}_{_{\small {\cal T}}}) }
    \leq \vertiii{ (\bsigma-\hat{\bsigma}, u - \hat{u}) }
      + \vertiii{(\hat{\bsigma}- \hat{\bsigma}_{_{\small {\cal T}}}, \hat{u} - \hat{u}_{_{\small {\cal T}}}) },
\end{equation}
where the first term represents the approximation error caused by the deep neural network and the second term is the numerical error by evaluating the FOSLS functional through numerical quadrature. How to estimate the former is still an open problem. The latter can be computed to
a desired accuracy through either uniform or adaptive partition of the $\Omega$, $\Gamma_D$, and $\Gamma_N$. A detailed algorithmic and theoretical discussions of the second term will be presented in a forthcoming paper. 

In (\ref{error}),  $(\hat{\bsigma}_{_{\small {\cal T}}}(x, \Theta),\,\hat{u}_{_{\small {\cal T}}}(x, \Theta))$ is assumed to be the exact solution of the minimization problem in (\ref{d-deep-fosls}). In practice,
problem (\ref{d-deep-fosls}) is solved numerically by an iterative method such as the method of (stochastic) gradient decent. Let $(\hat{\bsigma}^k_{_{\small {\cal T}}}(x, \Theta),\,\hat{u}^k_{_{\small {\cal T}}}(x, \Theta))$ be the algebraic approximation at the $k^{th}$ iterate, then the total error
of the discrete deep FOSLS method is bounded by the sum of the DNN approximation error,
the quadrature error, and the algebraic error as follows:
\begin{equation}\label{error+alg}
    \vertiii{ (\bsigma-\hat{\bsigma}^k_{_{\small {\cal T}}}, u - \hat{u}^k_{_{\small {\cal T}}}) }
    \leq \vertiii{ (\bsigma-\hat{\bsigma}, u - \hat{u}) }
      + \vertiii{(\hat{\bsigma}- \hat{\bsigma}_{_{\small {\cal T}}}, \hat{u} - \hat{u}_{_{\small {\cal T}}}) }
      + \vertiii{(\hat{\bsigma}_{_{\small {\cal T}}}- \hat{\bsigma}^k_{_{\small {\cal T}}}, \hat{u}_{_{\small {\cal T}}} - \hat{u}^k_{_{\small {\cal T}}}) }.
\end{equation}
Again, (\ref{error+alg}) is obtained by the triangle inequality.

\section{Numerical Experiments}

\begin{figure}[htbp]
\centering
\includegraphics[width=5.5in,height=1.17in]{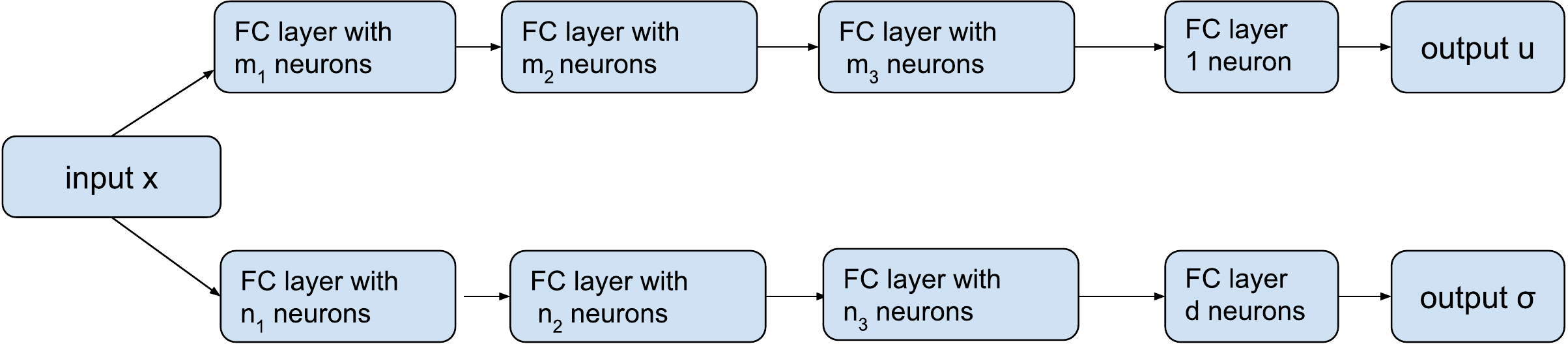}
\caption{Four-layer neural network for training $u(x)$ and $\bm{\sigma}(x)$. Each block consists of one fully-connected (FC) layer. $x$ is an arbitrary point in the domain $\Omega\subset\R^d$, and $m_l$ and $n_l$ are the respective numbers of neurons in the upper and lower branches at the $l^{\mbox{th}}$ layer.}
\label{network1}
\end{figure}

The solution $u(x)$ and the flux $\bm{\sigma}(x)$ in the FOSLS formulation are independent variables. This observation implies that an efficient DNN structure is to approximate them separately. Hence, a DNN to be employed consists of two branches: the upper and lower branches for the respective $u$ and $\bsigma$ (see Figure \ref{network1}). These two branches have no neuron connection. For numerical experiments in this paper, we use a four-layer neural network. Within each branch, a fully connected layer is implemented. 

Let $\Theta_u$ and $\Theta_{\bm{\sigma}}$ represent all parameters in the upper and lower branches, respectively. 
Denote by $\mathcal{N}_{u}^l$ and $\mathcal{N}_{\bm{\sigma}}^l$ the fully connected layer defined in (\ref{layerdef}) for the respective upper and lower branches. 
The four-layer neural network (see Figure \ref{network1}) defines two functions $u(x, \Theta_u)$ and $\bm{\sigma}(x,\Theta_{\bm{\sigma}})$ by the upper and lower branches:
\begin{eqnarray*}
u(x, \Theta_u)
= \mathcal{N}_{u}^4 \circ\mathcal{N}_{u}^3\circ \mathcal{N}_{u}^2\circ \mathcal{N}_{u}^1(x) 
\,\,\,\text{ and }\,\,\,\bm{\sigma}(x, \Theta_{\bm{\sigma}})
= \mathcal{N}_{\bm{\sigma}}^4\circ\mathcal{N}_{\bm{\sigma}}^3\circ \mathcal{N}_{\bm{\sigma}}^2\circ \mathcal{N}_{\bm{\sigma}}^1(x),
\end{eqnarray*}
respectively. Activation functions for the hidden and the output layers are usually different depending on the underlying application. In this paper, we use the same activation function for the hidden layers and identity for the output layer.
In the numerical experiments, both the leaky ReLU and sigmoid functions are tested for
the deep Ritz and the FOSLS methods, while the leaky ReLU activation function may not be used for the deep LS method as discussed in section~3.1.
Now, the deep FOSLS method is to find $(\bm{\sigma}(x,\Theta_{\bm{\sigma}}),\, u(x, \Theta_u))$
by minimizing the discrete FOSLS functional defined in (\ref{d-functiona11l}) over parameters $\Theta=(\Theta_{u}, \Theta_{\bm{\sigma}})$.
The deep LS and Ritz methods are to find $u(x, \Theta_u)$ (using only the upper branch)
by minimizing the corresponding discrete LS and energy functionals over parameters $\Theta_u$ (Remark 3.1).

To train (numerically compute) parameters $\Theta$ associated with the DNN functions $u(x,\,\Theta_u)$ and $\bsigma(x,\,\Theta_{\bm{\sigma}} )$, the Adam optimizer version of gradient descent \cite{kingma2015} is implemented as an iterative method to numerically solve the minimization problem in (\ref{d-deep-fosls}). The iterative parameter (may vary at each iteration) of the method of gradient decent is called the step size or learning rate.

Test problems in this section consist of a Poisson, a singularly perturbed reaction-diffusion equation, and 
an interface problem, all in one dimension.
As discussed in section 3.2, the FOSLS functional, similarly the energy and the LS functionals, are evaluated numerically based on a partition of the domain. For numerical results reported in sections~4.1, 4.2, and 4.3, we use a uniform partition of interval $[a,\,b]$: $a=x_0< x_1<
\cdots < x_n=b$ with $x_i=a+i\,h$ and $h=(b-a)/n$
for $i=0,\,1,\, ... ,\, n$.
Quadrature points in (\ref{d-functiona11l}) are chosen to be
the midpoints of subintervals: $x_{i-1/2}=a +h(2i-1)/2$ for $i=1,\,2,\, ... ,\, n$.
First-order derivative at midpoints in the functionals are approximated by the forward finite difference quotient, $\dfrac{v(x_{i-1/2})-v(x_{i-1/2}-\tau)}{\tau}$ with $\tau=h/2$.

 All experiments are replicated three times to reduce variability of random initialization of the method of gradient decent and the medians of three training results are reported. Numerical results are reported through the true error in the relative $L^2$ norm and the $H^1$ seminorm (or the energy norm) (see Tables \ref{poissonnumerical1}, \ref{singulartable}, and \ref{difftable}). Moreover, the exact solution vs numerical approximations are depicted in Figures \ref{poissonapp}, \ref{singularfig}, and \ref{difffig}. Note that only the figures for the FOSLS functional are presented as reference in Figures \ref{poissonapp} and \ref{singularfig} since results for the energy and the LS functionals are similar. For the deep FOSLS method, we also report numerical results on the approximation to the flux variable $\bsigma$ in the relative $L^2$ norm and the relative value of the FOSLS functional. A PyTorch implementation is released at \url{https://github.com/janiechen8/DeepLSMethod}.

\subsection{Poisson Equation}
The first test problem is a one-dimensional Poisson equation used in \cite{he2018relu}:
\begin{equation}\label{test1}
\left\{ \begin{array}{rclll}
 -u''(x)&=&f(x), \quad &x&\in \Omega=(0,\,1),\\[2mm]
 u&=&0, \quad &x&\in \partial \Omega = \{0,\,1\} 
 \end{array}\right.
 \end{equation}
with $f=-40000(x^3-2x^2/3+173x/1800+1/300)e^{-100(x-1/3)^2}$. Problem (\ref{test1}) has 
the following exact solution
\[
 u(x)=x\left(e^{-(x-\frac{1}{3})^2/0.01}-e^{-\frac{4}{9}/0.01}\right). 
 \]
A four-layer neural network ($m_1=n_1=24$ and $m_2=m_3=n_2=n_3=14$) with total 1246 parameters is implemented for the deep FOSLS method.

 The first numerical experiment is to show that with sufficient quadrature points for evaluating the FOSLS functional, accuracy of the deep FOSLS method is determined by the approximation property of the DNN structure (3.8). Denote $\bar{u}_{\tau}$ and $\bar{\bsigma}_{\tau}$ as the network outputs of $u$ and $\bsigma$, respectively. Using the leaky ReLU activation function, a fixed learning rate of $0.0005$ and 10000 iterations, Table \ref{meshsize} shows that $800$ quadrature points are enough to accurately evaluate the FOSLS functional.

\begin{figure}[htbp]
  \centering 
   \subfigure[FOSLS $u$ with Sigmoid activation]{ 
    \label{fig:subfig:cc} 
    \includegraphics[width=2.2in]{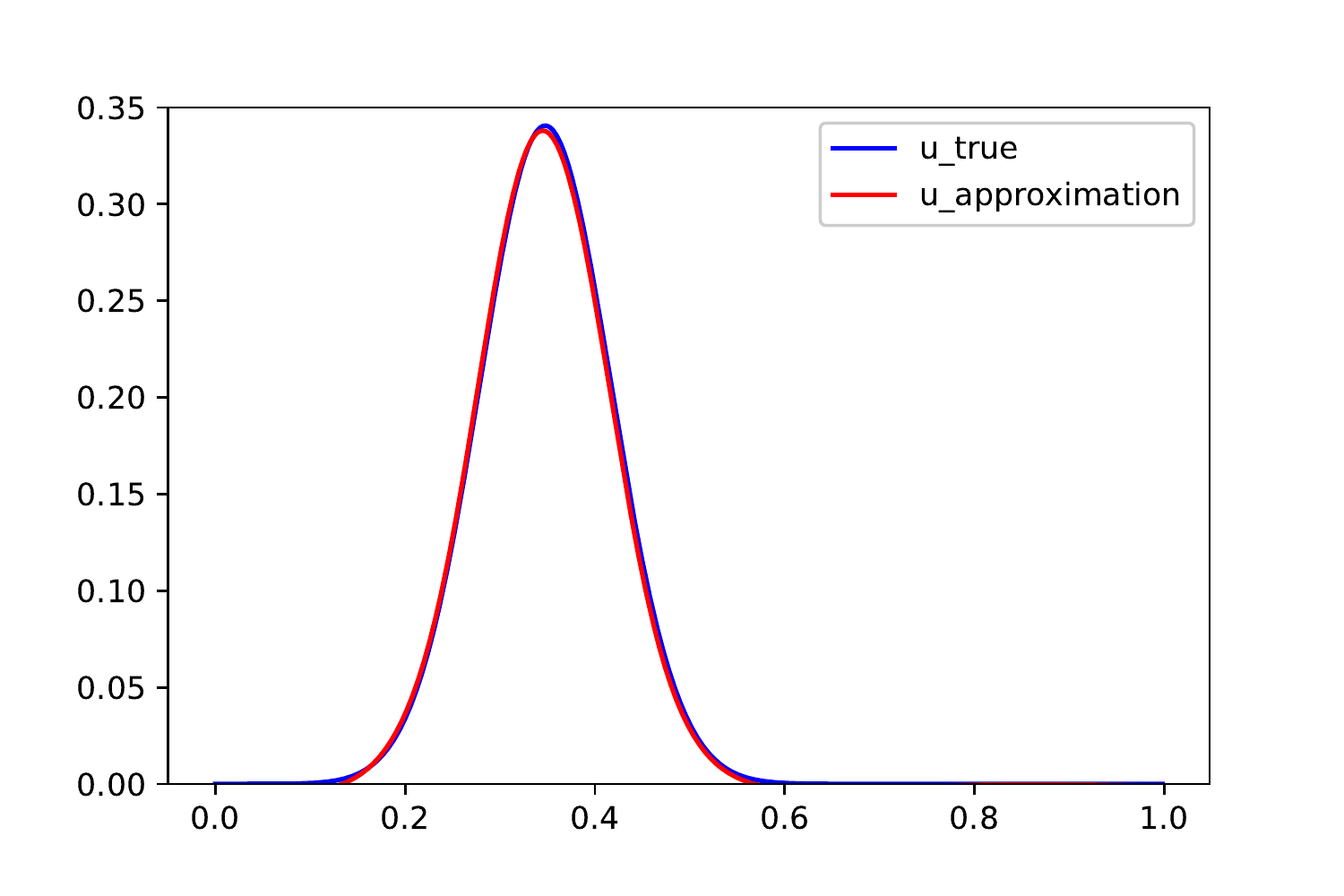}} 
  \hspace{0.3in} 
  \subfigure[FOSLS $\bsigma$ with Sigmoid activation]{ 
    \label{fig:subfig:dd} 
    \includegraphics[width=2.2in]{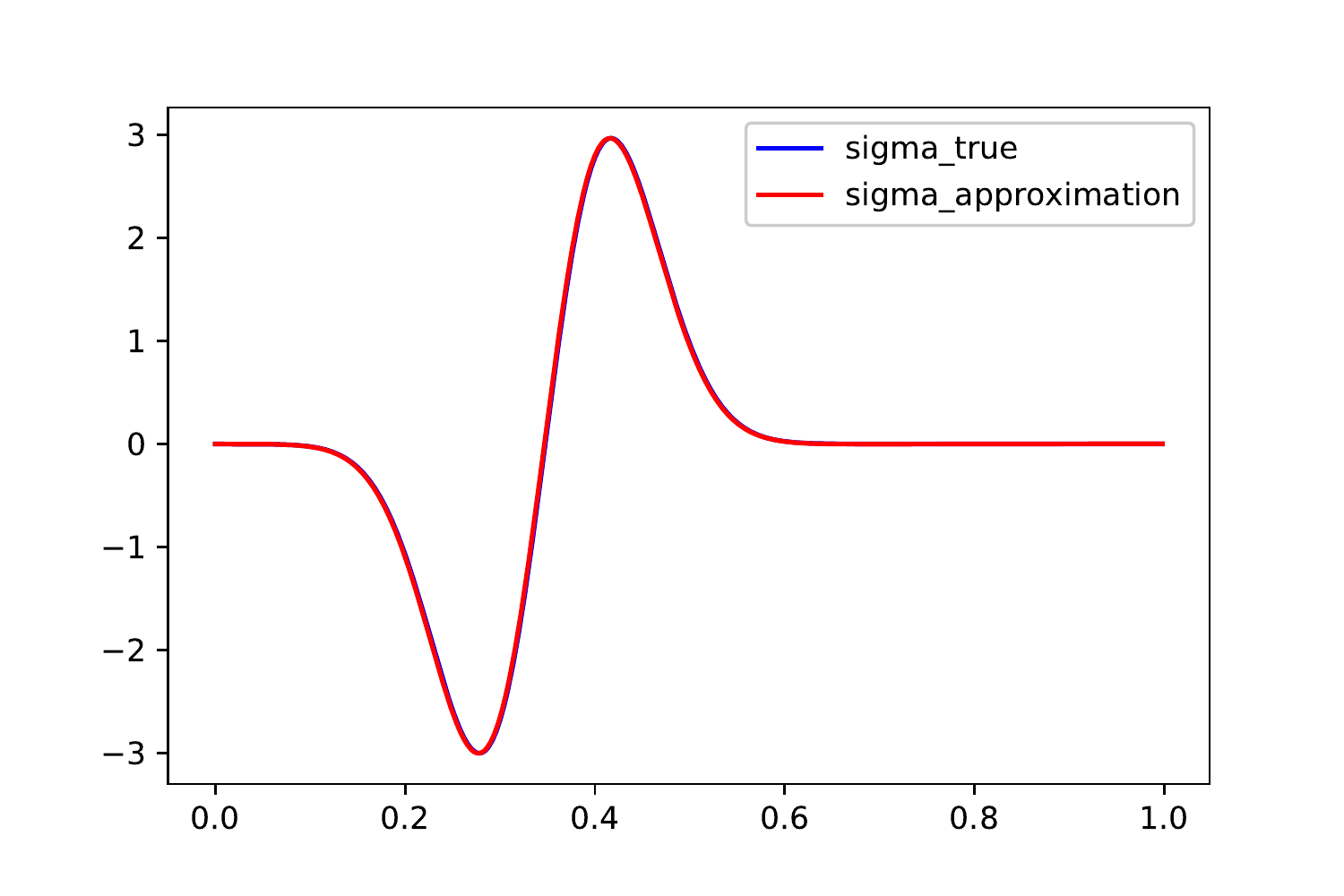}} 
     \hspace{0.3in} 
  \caption{Poisson equation approximation results with FOSLS functional and sigmoid activation}
  \label{poissonapp} 
\end{figure}

 The goal of the second numerical experiment is to report numerical performances when using different functionals as well as activation functions. With the same learning rate and iteration number, Table~\ref{poissonnumerical1} and Figure \ref{poissonapp} show that all three methods are able to accurately approximate the solution of the 
Poisson equation. Due to smoothness of the exact solution, the deep LS method performs slightly better than the other two methods; moreover, the sigmoid function is more accurate than the leaky ReLU function possibly because of exponential feature of the exact solution.

\begin{table}[ht]
\caption{Relative errors of Poisson equation with different number of quadrature points}\label{meshsize}
\centering
\begin{tabular}{  |l |c | c | c | c| }
	\hline
	\diagbox[width=11.9em, trim=l, dir=SE]{quadrature points}{Relative errors}  &  $\dfrac{\|u-\bar{u}_{\tau}\|_0}{\|u\|_0}$ &  
 $\dfrac{|u-\bar{u}_{\tau}|_1}{|u|_1}$& 
	$\dfrac{\|\bm{\sigma}-\bar{\bsigma}_{\tau}\|_{0}}{\|\bm{\sigma}\|_{0}}$ &
	$\dfrac{ G^{1/2}(\bar{\bsigma}_{\tau},\,\bar{u}_{\tau};{\bf f})}{\vertiii{(\bm{\sigma},\,u)}}$ \\[4mm] \hline
	200 & 0.065238  & 0.109056   
	& 0.056508 &0.098030 \\ \hline 
	400 &  0.048421 & 0.167703   
	& 0.026564  &0.095498\\ \hline 
	800 & 0.025238  & 0.106552   
	&0.020481 &0.068702  \\ \hline 
	1600 & 0.024631  & 0.114932   
	&0.020091  &0.063403   \\ \hline 
\end{tabular}
\end{table}

\begin{table}[ht]
\caption{Relative errors of Poisson equation with different functionals, activation functions and quadrature points}\label{poissonnumerical1}
\centering
\begin{tabular}{  |l |c | c | c |  c| c }
	\hline
	\diagbox[width=13em, trim=l, dir=SE]{Loss and activation}{Relative errors}
  &  $\dfrac{\|u-\bar{u}_{\tau}\|_0}{\|u\|_0}$ &  
 $\dfrac{|u-\bar{u}_{\tau}|_1}{|u|_1}$& 
	$\dfrac{\|\bm{\sigma}-\bar{\bsigma}_{\tau}\|_{0}}{\|\bm{\sigma}\|_{0}}$ &
	$\dfrac{ G^{1/2}(\bar{\bsigma}_{\tau},\,\bar{u}_{\tau};{\bf f})}{\vertiii{(\bm{\sigma},\,u)}}$ \\[4mm] \hline
	Energy (LeakyReLU \& 800 points) &  0.029161 & 0.160666   & ---& ---  \\ \hline 
	FOSLS (LeakyReLU \& 800 points) & 0.025238  & 0.106552  & 0.020481 & 0.068702 \\ \hline 
	Energy (Sigmoid \& 200 points) &  0.013144 & 0.026246   & ---& ---  \\ \hline
	LS (Sigmoid \& 200 points) &  0.008876 & 0.009108    & ---& ---  \\ \hline
	FOSLS (Sigmoid \& 200 points) & 0.013505  & 0.019830  & 0.008897 & 0.045650  \\ \hline
\end{tabular}
\end{table}

\subsection{Singularly Perturbed Reaction-Diffusion Equation}
The second test problem is a singularly perturbed reaction-diffusion equation:
\begin{equation}\label{test2}
\left\{ \begin{array}{rclll}
  -\varepsilon^2  u''(x) + u(x) &=& f(x), \quad &x&\in \Omega = (-1,\,1),\\[2mm]
 u&=&0, \quad &x&\in \partial \Omega = \{-1,\,1\}.
 \end{array}\right.
 \end{equation}
For $f=-2\left(\varepsilon-4x^2\text{tanh}(\frac{1}{\varepsilon}(x^2 - \frac{1}{4}))\right)\left(1/\text{cosh}(\frac{1}{\varepsilon}(x^2 - \frac{1}{4}))\right)^2 + \tanh(\frac{1}{\varepsilon}(x^2 - \frac{1}{4})) - \tanh(\frac{3}{4\varepsilon}) $, problem (\ref{test2}) has the following exact solution 
\[u(x) = \tanh\left(\frac{1}{\varepsilon}(x^2 - \frac{1}{4})\right) - \tanh\left(\frac{3}{4\varepsilon}\right).
\]
With $\bm{\sigma} = - \varepsilon^2 u'$, 
the corresponding FOSLS functional defined in (\ref{fosls1a}) is of the form
\begin{eqnarray*}
\mathcal{G}(\bm{\tau},v;f)
&=& \|\bm{\tau}^\prime + v -f\|^2_{0,\Omega} + \big\| \bm{\tau} / \epsilon  + \epsilon v^\prime\big\|^2_{0,\Omega} + \alpha\,\|u\|^2_{1/2,\partial \Omega},
\end{eqnarray*}
and the corresponding energy norms are $\vertiii{(\btau,\,v)}=\left(\vertiii{\btau}^2 + \vertiii{v}^2\right)^{1/2}$ with
\[
\vertiii{v}=\left( \|v\|^2_{0,\Omega} + \|\epsilon v^\prime\|^2_{0,\Omega}
\right)^{1/2}
\quad\mbox{and}\quad
\vertiii{\btau}=\left(\|\btau/\epsilon\|^2_{0,\Omega} + \|\btau^\prime\|^2_{0,\Omega}
\right)^{1/2}.
\]

The goal of this numerical experiment is to test the performance of deep learning based method for problems with boundary and/or interior layers which pose difficulty for mesh-based methods such as finite element, finite difference, etc.
The four-layer neural network depicted in Figure~\ref{network1} is implemented with the following setting: $m_1=n_1=32$ and $m_2=m_3=n_2=n_3=24$. This network has 2962 parameters. Uniformly distributed $2000$ quadrature points are used for evaluating different cost functionals. The learning rate starts with 0.001, and is reduced by half for every $5000$ iterations. This learning rate decay strategy is adopted for accelerating the training (iterative) process.
 
\begin{figure}[htbp]
  \centering 
    \subfigure[FOSLS $u$ with Leaky ReLU]{ 
    \label{fig:singular:a} 
    \includegraphics[width=2.2in]{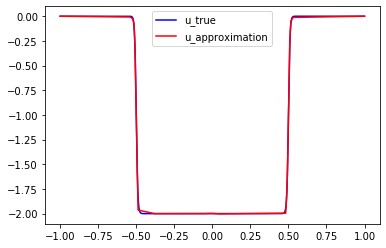}} 
  \hspace{0.5in} 
  \subfigure[FOSLS $\bsigma$ with Leaky ReLU]{ 
    \label{fig:singular:b} 
    \includegraphics[width=2.2in]{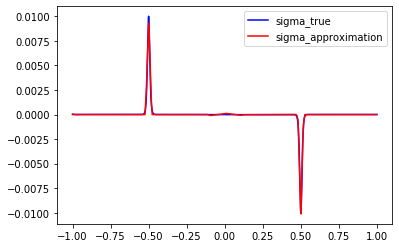}}
    \hspace{0.5in}  
  \caption{Singularly perturbed equation: approximation results with FOSLS functional and Leaky ReLU} 
  \label{singularfig} 
\end{figure}

For $\varepsilon=0.01$ and $\alpha=1$, after 20000 iterations, the median results are reported in Table \ref{singulartable} and Figure \ref{singularfig}. All three methods exhibit accurate approximation to the solution with interior layers. For both the leaky ReLU and sigmoid activation functions, the deep FOSLS method is more accurate than the deep Ritz method. Again, the DNN using the sigmoid function is more accurate than that using the leaky ReLU function, possibly due to exponential feature of the exact solution.

An interesting observation from Figure 3 is that the DNN-based methods do not produce  overshooting and oscillations, unlike
mesh-based traditional numerical methods without strategies such as limiter, etc. This could indicate that the deep FOSLS, LS, and Ritz methods have potential to accurately approximate problems with boundary and/or interior layers.

\begin{table}[ht]
\caption{Relative errors of singularly perturbed equation with different loss and activation functions}\label{singulartable}
\centering
\begin{tabular}{  |l |c | c | c |  c| c }
\hline
	\diagbox[width=13em, trim=l, dir=SE]{Loss and activation}{Relative errors}  &  $\dfrac{\|u-\bar{u}_{\tau}\|_0}{\|u\|_0}$ &  
 $\dfrac{\vertiii{u-\bar{u}_{\tau}}}{\vertiii{u}}$& 
	$\dfrac{\|\bm{\sigma}-\bar{\bsigma}_{\tau}\|_{0}}{\|\bm{\sigma}\|_{0}}$ &
	$\dfrac{ G^{1/2}(\bar{\bsigma}_{\tau},\,\bar{u}_{\tau};{\bf f})}{\vertiii{(\bm{\sigma},\,u)}}$ \\[4mm]\hline
    Energy functional (LeakyReLU) &  0.011316& 0.026179   & ---& ---  \\ \hline 	
	FOSLS functional (LeakyReLU) & 0.006654  & 0.020810  & 0.099863 & 0.031482 \\ \hline 
	Energy functional (Sigmoid) &  0.003019 & 0.004612  & ---& ---  \\ \hline 
	LS functional (Sigmoid) &  0.000910 & 0.002088   & ---& ---  \\ \hline 
	FOSLS functional (Sigmoid) & 0.001403  & 0.001711  & 0.211490 & 0.014825 \\ \hline
\end{tabular}
\end{table}

\subsection{Interface Problem}
To test the performance of three cost functionals for non-smooth problems, we experimented a one-dimensional interface equation as follows.
\begin{equation}\label{test33}
\left\{ \begin{array}{rclll}
  -\left(a  u^\prime(x)\right)^\prime  &=& f(x), \quad &x&\in \Omega = (0,\,1),\\[2mm]
 u&=&0, \quad &x&\in \partial \Omega = \{0,\,1\},
 \end{array}\right.
 \end{equation}
where $a = 1$ for $x\in  (0,\, \frac{1}{2})$ and $a = k$ for $x\in  (\frac{1}{2},\, 1)$. 
It is well-known that solutions of interface problems are not smooth, in particular, not in $H^2(\Omega)$.
For 
\begin{equation*}
f(x)=\left\{ \begin{array}{rclll}
  8k(3x-1), \quad &x&\in  (0,\,\frac{1}{2}),\\[2mm]
  4k(k+1), \quad &x&\in  (\frac{1}{2},\,1),
 \end{array}\right.
 \end{equation*}
problem (\ref{test33}) has the following exact solution 
\begin{equation*}\label{test3ff}
u(x)=\left\{ \begin{array}{rclll}
  4kx^2(1-x), \quad &x&\in  (0,\,\frac{1}{2}),\\[2mm]
  [2(k+1)x-1](1-x), \quad &x&\in  (\frac{1}{2},\,1).
 \end{array}\right.
 \end{equation*}
 Note that derivative of the true solution is discontinuous  at point $x=0.5$.
With $\bm{\sigma} = - a u'$, the corresponding FOSLS functional defined in (\ref{fos}) has the form
\begin{eqnarray*}
\mathcal{G}(\bm{\tau},v;f)
&=& \|\bm{\tau}^\prime -f\|^2_{0,\Omega} + \big\| a^{-1/2}\bm{\tau}   +  a^{1/2}v^\prime\big\|^2_{0,\Omega} + \alpha\,\|u\|^2_{1/2,\partial \Omega}.
\end{eqnarray*}

The same network structure is implemented as the one used in section 4.2.
Numerical evaluations of the functionals are done on a uniform partition of the interval $[0,1]$ with $h=0.002$. A same learning rate decay strategy is adopted here as described in section 4.2.
 
\begin{figure}[htbp]
  \centering 
    \subfigure[Energy Variation $u$]{ 
    \label{fig:diff:a} 
    \includegraphics[width=2.2in]{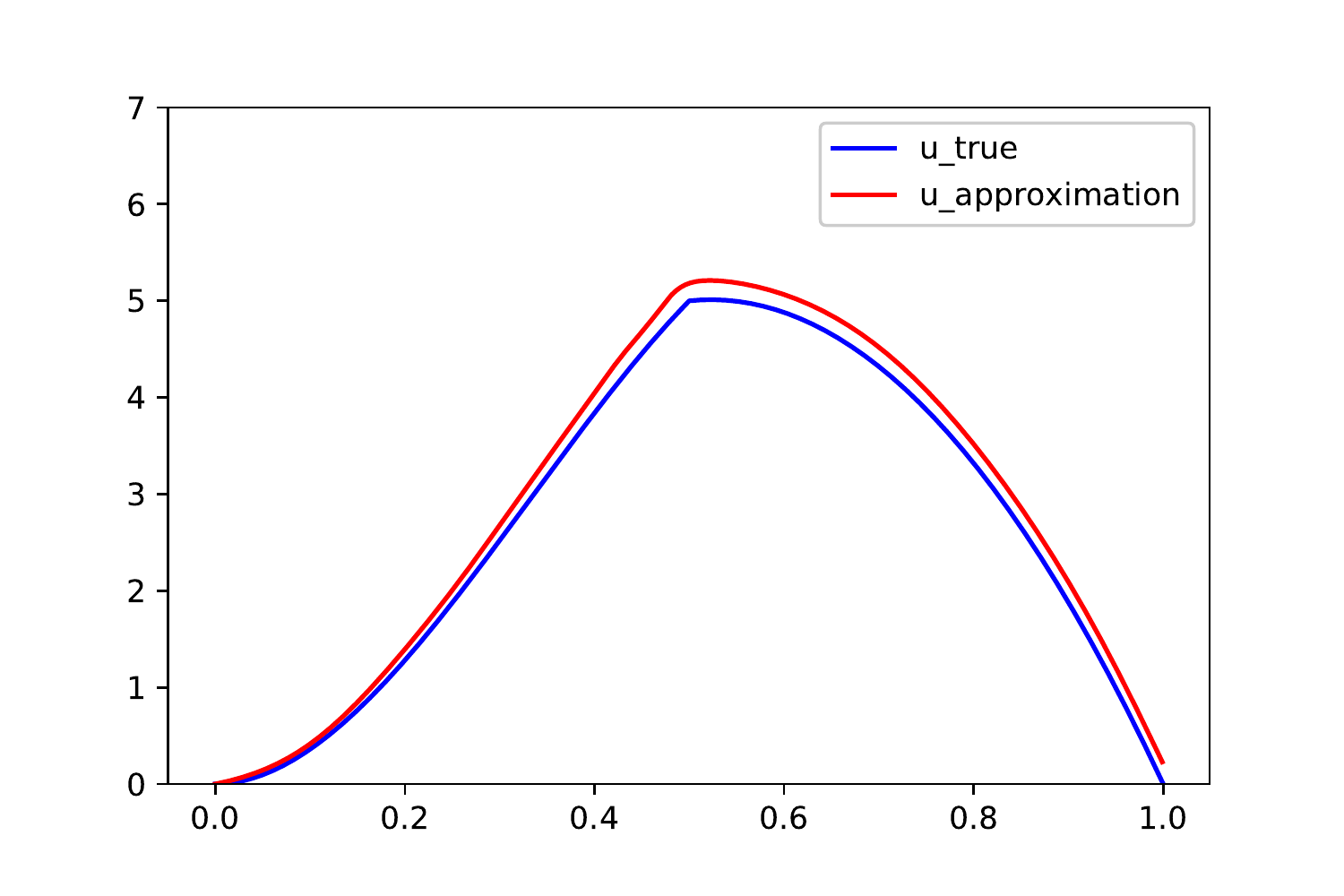}} 
  \hspace{0.5in} 
  \subfigure[Energy Variation $-\alpha u'$]{ 
  \label{fig:diff:a2} 
    \includegraphics[width=2.2in]{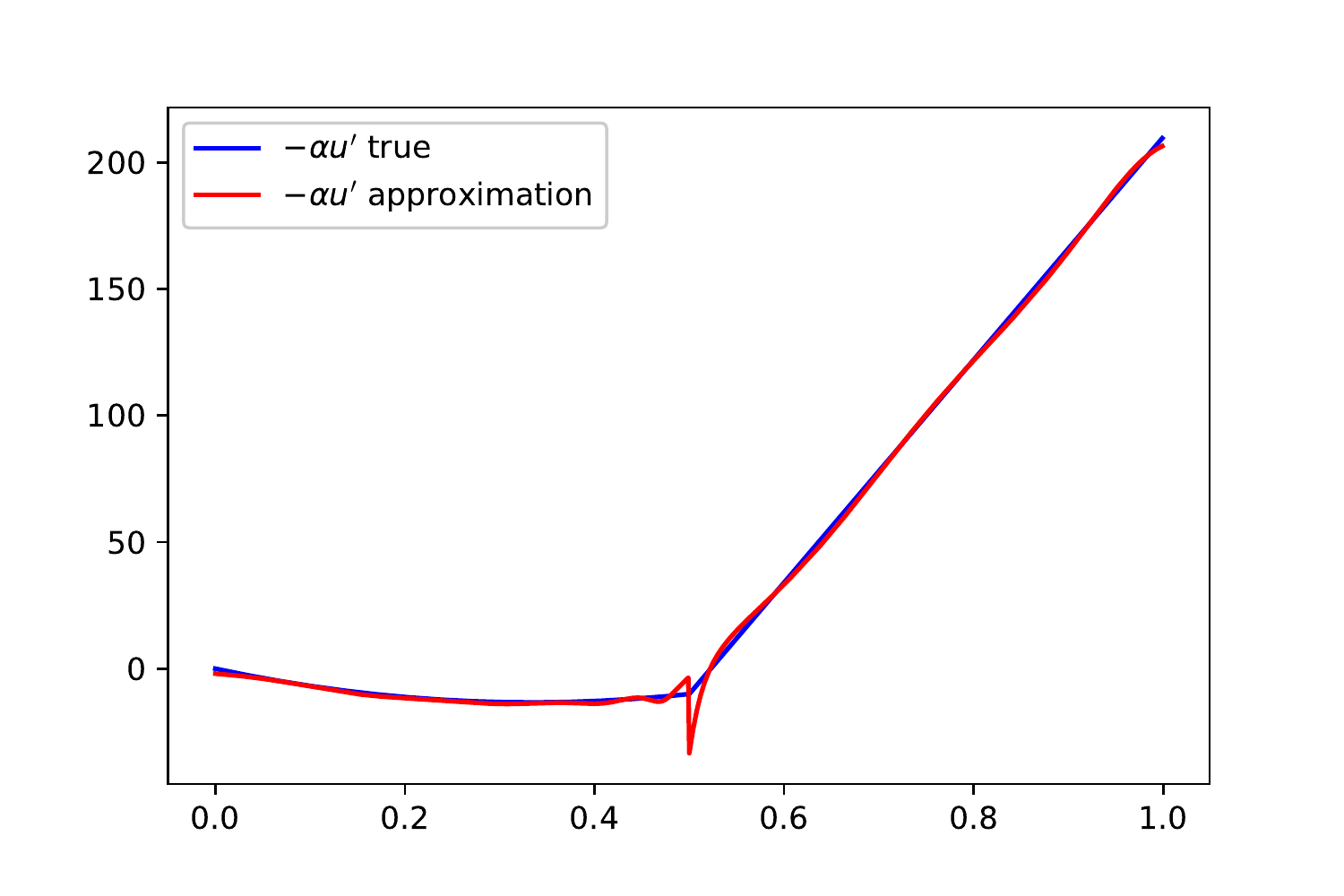}} 
  \hspace{0.5in} 
  \subfigure[LS $u$]{ 
    \label{fig:diff:b} 
    \includegraphics[width=2.2in]{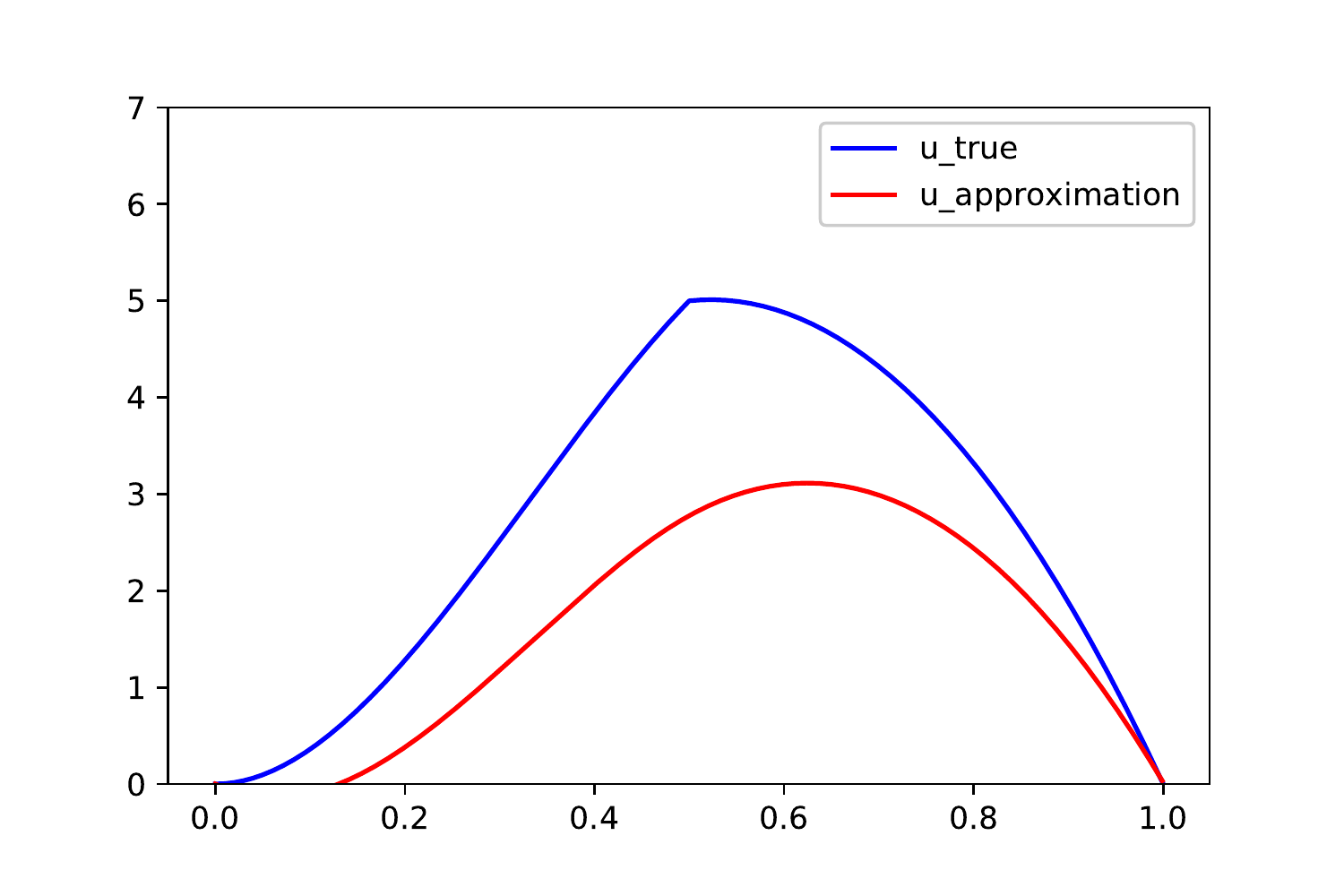}}
     \hspace{0.5in} 
\subfigure[LS $-\alpha u'$]{ 
  \label{fig:diff:b2} 
    \includegraphics[width=2.2in]{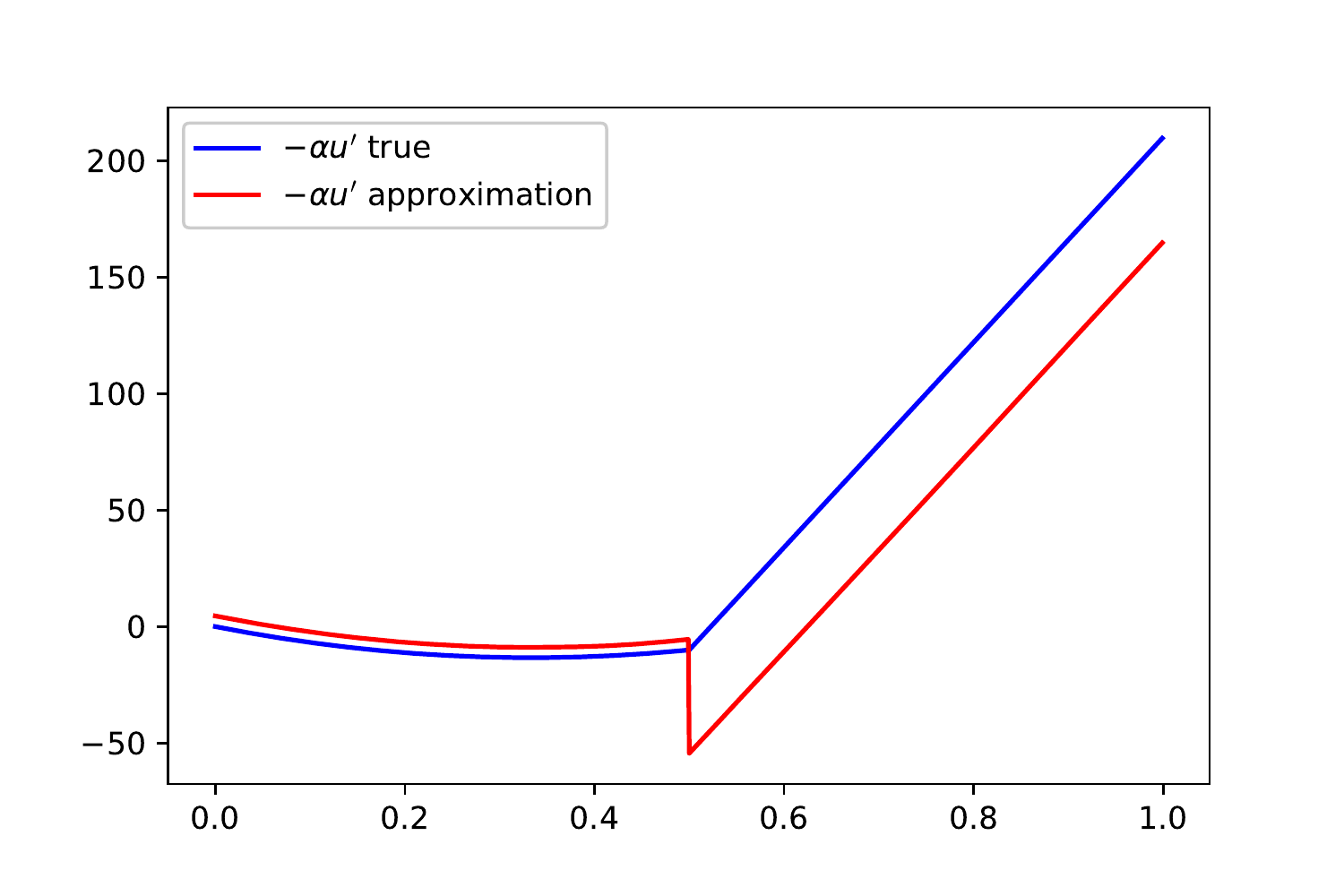}} 
    \hspace{0.5in} 
    \subfigure[FOSLS $u$]{ 
    \label{fig:diff:c} 
    \includegraphics[width=2.20in]{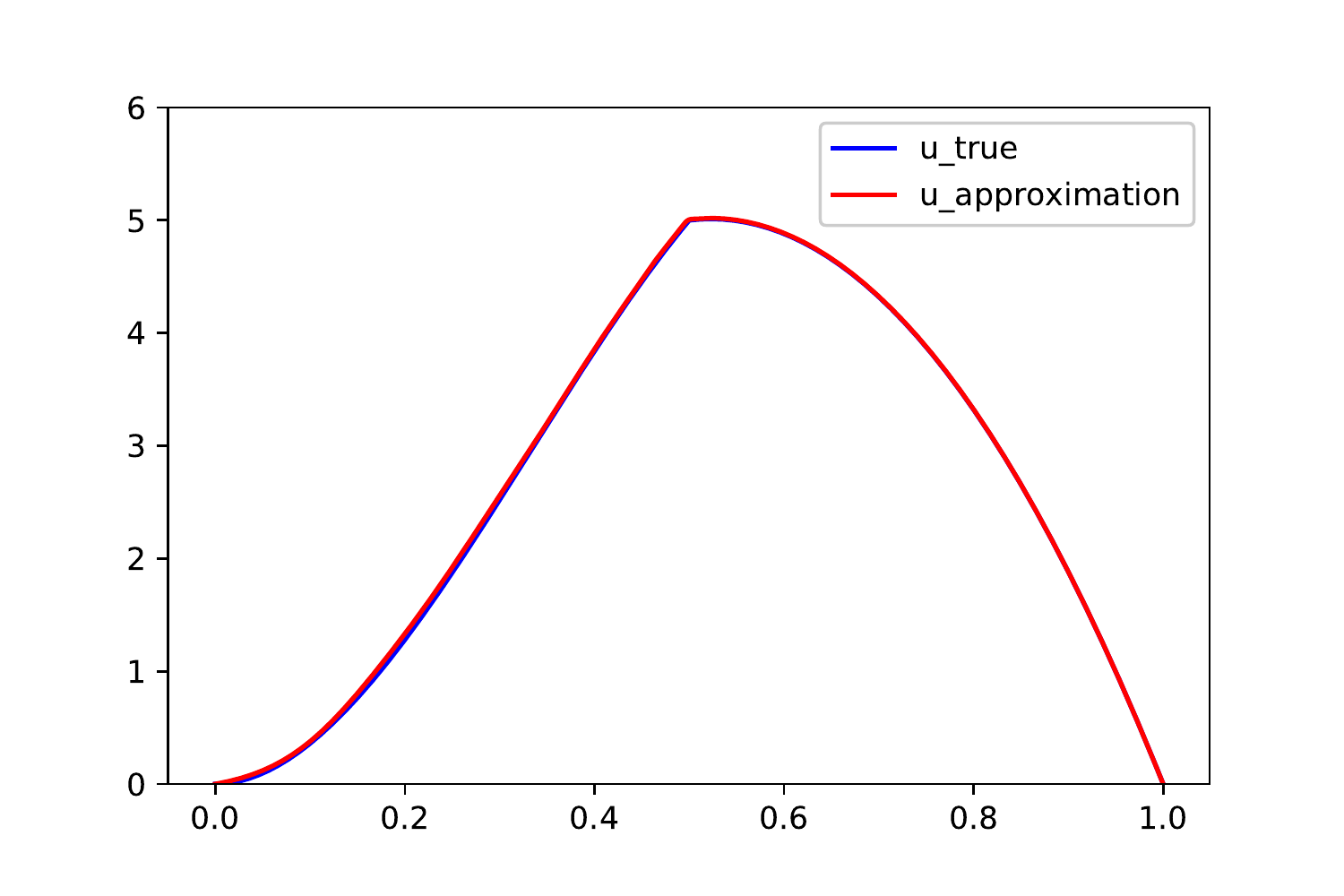}} 
    \hspace{0.5in} 
    \subfigure[FOSLS $\bm{\sigma}$]{ 
    \label{fig:diff:d} 
    \includegraphics[width=2.20in]{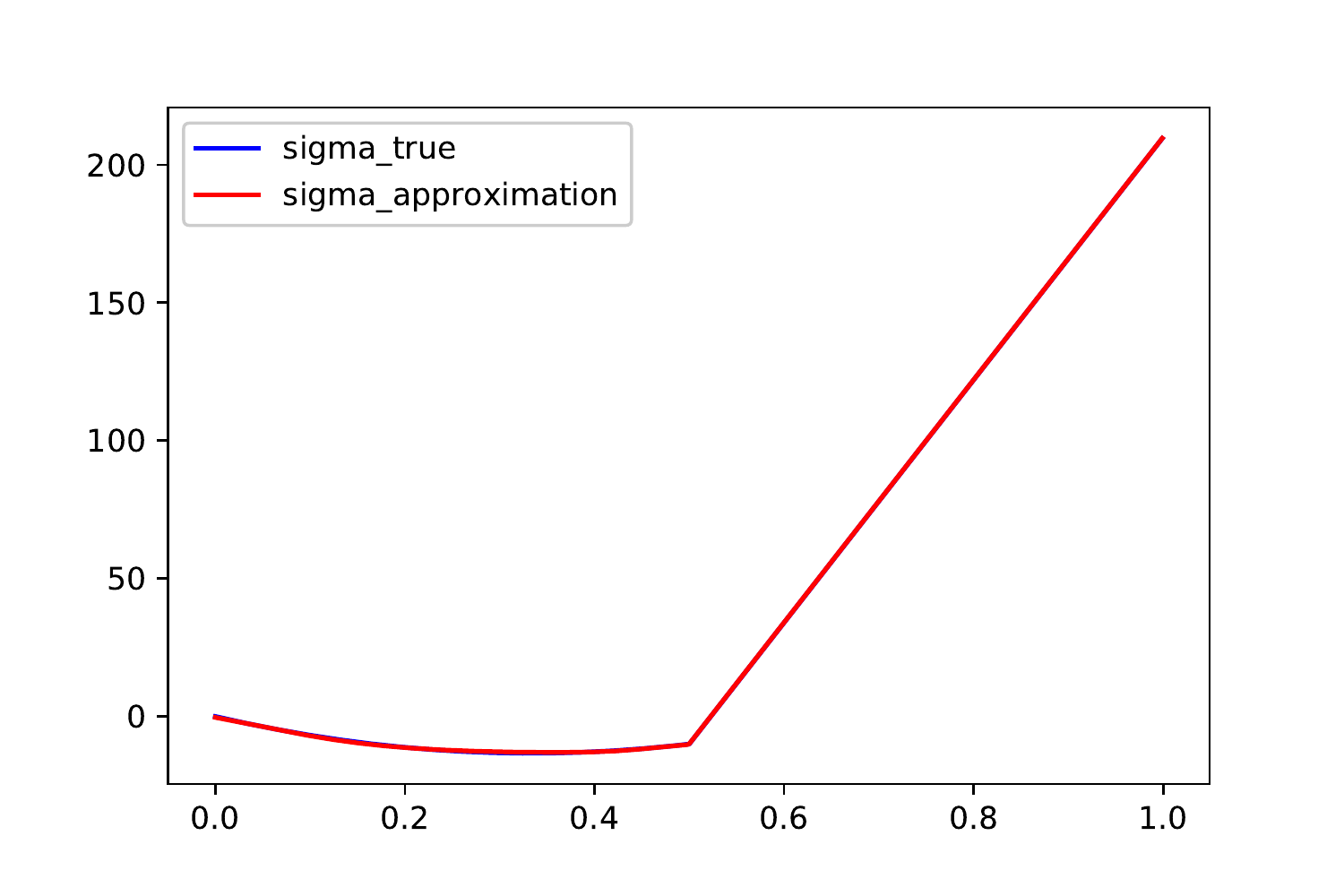}} 
  \caption{Interface problem approximation results using different loss functions (all with Sigmoid activation function)} 
  \label{difffig} 
\end{figure}

For $k=10$ and $\alpha=1$, the numerical result after 20000 iterations are reported in Table \ref{difftable} and Figure~\ref{difffig}. The results show that the deep FOSLS method is significantly better than the deep Ritz method, while the deep LS method fails to approximate the solution well. This verifies Remark 2.1, i.e., the deep 
LS method is only applicable to sufficiently smooth problems. Moreover, since the true solution of this problem is a piecewise polynomial, as expected that the leaky ReLU activation function gives a better performance than the sigmoid function. This indicates that the choice of activation function is problem dependent, and we may use the relative value of the FOSLS functional to guide this choice in real-world applications where the true solutions are unknown.

\begin{table}[ht]
\caption{Relative errors of interface problem with different loss and activation functions}\label{difftable}
\centering
\begin{tabular}{ |l |c | c |  c| c }
\hline
	\diagbox[width=13em, trim=l, dir=SE]{Loss function }{Relative errors}  &  $\dfrac{\|u-\bar{u}_{\tau}\|_0}{\|u\|_0}$ &  
	$\dfrac{\|\bm{\sigma}-\bar{\bsigma}_{\tau}\|_{0}}{\|\bm{\sigma}\|_{0}}$ &
	$\dfrac{ G^{1/2}(\bar{\bsigma}_{\tau},\,\bar{u}_{\tau};{\bf f})}{\vertiii{(\bm{\sigma},\,u)}}$ \\[4mm] \hline
	Energy functional (Sigmoid) &  0.054705  & ---& ---  \\ \hline 
	LS functional (Sigmoid) &  0.397965 &  ---& ---  \\ \hline 
	FOSLS functional (Sigmoid) & 0.007137   & 0.001870 & 0.005073 \\ \hline
	Energy functional (Leaky ReLU)  & 0.041087 & ---& ---  \\ \hline 
	FOSLS functional (Leaky ReLU) & 0.002840  &  0.000686 & 0.001406 \\ \hline	
\end{tabular}
\end{table}

\subsection{Adaptive Numerical Quadrature}

Numerical results reported in the previous sections employed uniform quadrature points. As discussed in section 1, one appealing feature of FOSLS function is that the value of the corresponding FOSLS functional is an accurate a posteriori error estimator which can be used to guide an adaptive control of the quadrature points selection. In this section, we report numerical results of the deep FOSLS method with the leaky ReLU using local and global refined partitions 
for the test problem in section 4.1. The same network structure and learning rate as those in section 4.1 are used.

To this end, we first describe adaptive numerical quadrature. Let $\mathcal{T}^{old}$ be the current partition of the domain $\Omega$. For each subdomain $K\in\mathcal{T}^{old}$, let $x_K\in K$ be the quadrature point (e.g., the centroid of $K$). Let $\left(\bm{\sigma}(x, \Theta_{\bm{\sigma}}), u(x, \Theta_u)\right)$ be the deep FOSLS approximation associated with the current partition $\mathcal{T}^{old}$. If the relative value of the FOSLS functional at $\left(\bm{\sigma}(x, \Theta_{\bm{\sigma}}), u(x, \Theta_u)\right)$ is not within the prescribed tolerance, we create a new partition $\mathcal{T}^{new}$ by refining the old partition $\mathcal{T}^{old}$ as follows:
\begin{itemize}
\item for each $K\in\mathcal{T}^{old}$, compute local indicator
\[
   \eta(x_K)= \left( \big(\mbox{div}\, \bm{\sigma}+X u -f\big)^2 (x_{\small K}, \Theta)
    +\big( A^{-1/2} \bm{\sigma} + A^{1/2}\nabla u  \big)^2 (x_{\small K}, \Theta)\right) |K|,
   \] 
\item refine subdomain $K\in \mathcal{T}^{old}$
if $\eta(x_K)$ is among the top $10\%$ of the largest indicators.
\end{itemize}
A subdomain may be refined, e.g., by bisection in low dimensions or by some aggressive refinements in high dimensions.

Starting with a uniform partition of interval $[0,\,1]$ with $h=0.005$,
Table 5 reports relative values of the FOSLS functional at the current approximations on both local and global refined, and uniformly distributed partitions. All three methods used a total of 10000 iterations. The local refinement method refines the quadrature points adaptively at every 2000 iterations, and global refinement method refines only once after 5000 iterations.
Clearly, Table 5 shows that locally refined partition is better than globally uniform partitions. 




\begin{table}[ht]
\caption{Comparison of locally refined and uniform partitions}\label{adaptive_table}
\centering
\begin{tabular}{  |l |c|  }
\hline
	\diagbox[width=13em, trim=l, dir=SE]{Methods }{Relative errors}  & 
	$\dfrac{ G^{1/2}(\bar{\bsigma}_{\tau},\,\bar{u}_{\tau};{\bf f})}{\vertiii{(\bar{\bsigma}_{\tau},\,\bar{u}_{\tau})}}$ \\[4mm] \hline
	Local refinement  of $200$ to $292$ quadrature points   & 0.085691 \\ \hline
		Global refinement of $200$ to $400$ quadrature points & 0.100553  \\ \hline 
	Uniform distribution of $292$ quadrature points   & 0.102849 \\ \hline 
\end{tabular}
\end{table}

\section{Discussion and Conclusion}
 
We proposed the deep FOSLS method by using DNNs to approximate solutions of PDEs
and modified the deep Ritz and the deep LS methods by treating boundary conditions in a balance way. 
While the deep Ritz and LS methods are applicable to problems having underlying minimization principle
and smooth problems, respectively,
the deep FOSLS method is applicable to
a much larger class of problems.

Both the deep LS and FOSLS methods are based on the least-squares principle applied to the respective original PDEs and a first-order system of the original PDEs. A striking feature of the least-squares principle is that values of the LS and FOSLS functionals provide feedback for automatically controlling numerical processes such as the numbers of neurons and layers in DNN, the number and the location of quadrature points for evaluating the functionals. Adaptive control first on numerical evaluation of the least-squares functionals (see preliminary numerical results in section 4.4) and then on DNN structure will be topics of our further study on the deep least-squares methods. Finally,  
unlike finite elements, DNN generates function in $H^2(\Omega)$ when using smooth activation functions. This means that the deep LS method is a competitive method for smooth problems.

With limited knowledge on approximation theory of DNNs, in order to accurately evaluate the functionals, inequality
(\ref{error}) and similar inequalities
in the $H^1$ and $H^2$ norms for the respective deep Ritz and LS methods shed some lights on how to adaptively choose quadrature points for a fixed DNN structure. Similarly, inequality (\ref{error+alg}) plus an algebraic error estimator provides a guidance on when to terminate the iterative process.

Comparing with traditional mesh-based numerical methods such as finite difference, finite volume, and finite element, etc.,
DNN provides a new class of functions that is meshless and ``pointless'' and that has the attractive feature of the moving mesh method. This explains why the deep FOSLS, LS, and Ritz methods approximate well the singularly perturbed reaction diffusion equation with a sharp interior layer (see section 4.2); in particular, the DNN approximations exhibit no overshooting and no oscillation which are common numerical defects for mesh-based traditional numerical methods without strategies such as limiter, etc.

\bigskip
\bibliographystyle{ieee}
\bibliography{Reference}

\end{document}